\documentclass[runningheads]{llncs}
\usepackage{multirow}
\usepackage{multicol}
\usepackage{amsmath,amssymb,amsfonts}
\usepackage{graphicx}
\usepackage{textcomp}
\usepackage{wrapfig}
\usepackage{caption}
\usepackage{subcaption}
\usepackage{float}
\usepackage{xcolor}
\usepackage{adjustbox}

\begin{document}
\title{Wine Characterisation with Spectral Information and Predictive Artificial Intelligence}
\titlerunning{Wine Characterisation with Spectral Information and Predictive AI}
%
\author{Jianping Yao\inst{1} \and
Son N. Tran\inst{1} 
\and
Hieu Nguyen\inst{1} \and
Samantha Sawyer \inst{2}
\and
Rocco Longo \inst{2,3}
}
\authorrunning{J. Yao et al.}
%
\institute{School of Information and Technology, University of Tasmania, TAS 7249, Australia
\email{\{jianping.yao, sn.tran, hieu.nguyen\}@utas.edu.au}
\and
Tasmania Institute of Agriculture. Mt Pleasant, TAS 7250, Australia
\email{samantha.sawyer@utas.edu.au}\\
\and
Winequip, Dudley Park, Adelaide, SA 5008, Australia
\email{rocco@winequip.com.au}
}
\maketitle              
\begin{abstract}
The purpose of this paper is to use absorbance data obtained by human tasting and an ultraviolet-visible (UV-Vis) scanning spectrophotometer to predict the attributes of grape juice (GJ) and to classify the wine's origin, respectively. The approach combined machine learning (ML) techniques with spectroscopy to find a relatively simple way to apply them in two stages of winemaking and help improve the traditional wine analysis methods regarding sensory data and wine's origins. This new technique has overcome the disadvantages of the complex sensors by taking advantage of spectral fingerprinting technology and forming a comprehensive study of the employment of AI in the wine analysis domain. In the results, Support Vector Machine (SVM) was the most efficient and robust in both attributes and origin prediction tasks. Both the accuracy and F1 score of the origin prediction exceed 91\%. The feature ranking approach found that the more influential wavelengths usually appear at the lower end of the scan range, 250 nm (nanometers) to 420 nm, which is believed to be of great help for selecting appropriate validation methods and sensors to extract wine data in future research. The knowledge of this research provides new ideas and early solutions for the wine industry or other beverage industries to integrate big data and IoT in the future, which significantly promotes the development of  'Smart Wineries'.

\keywords{Smart agriculture \and Machine learning \and Wine characterisation}
\end{abstract}
\section{Introduction}
\begin{figure}[ht]
\vskip -0.5cm
\centering
\includegraphics[width=0.85\textwidth]{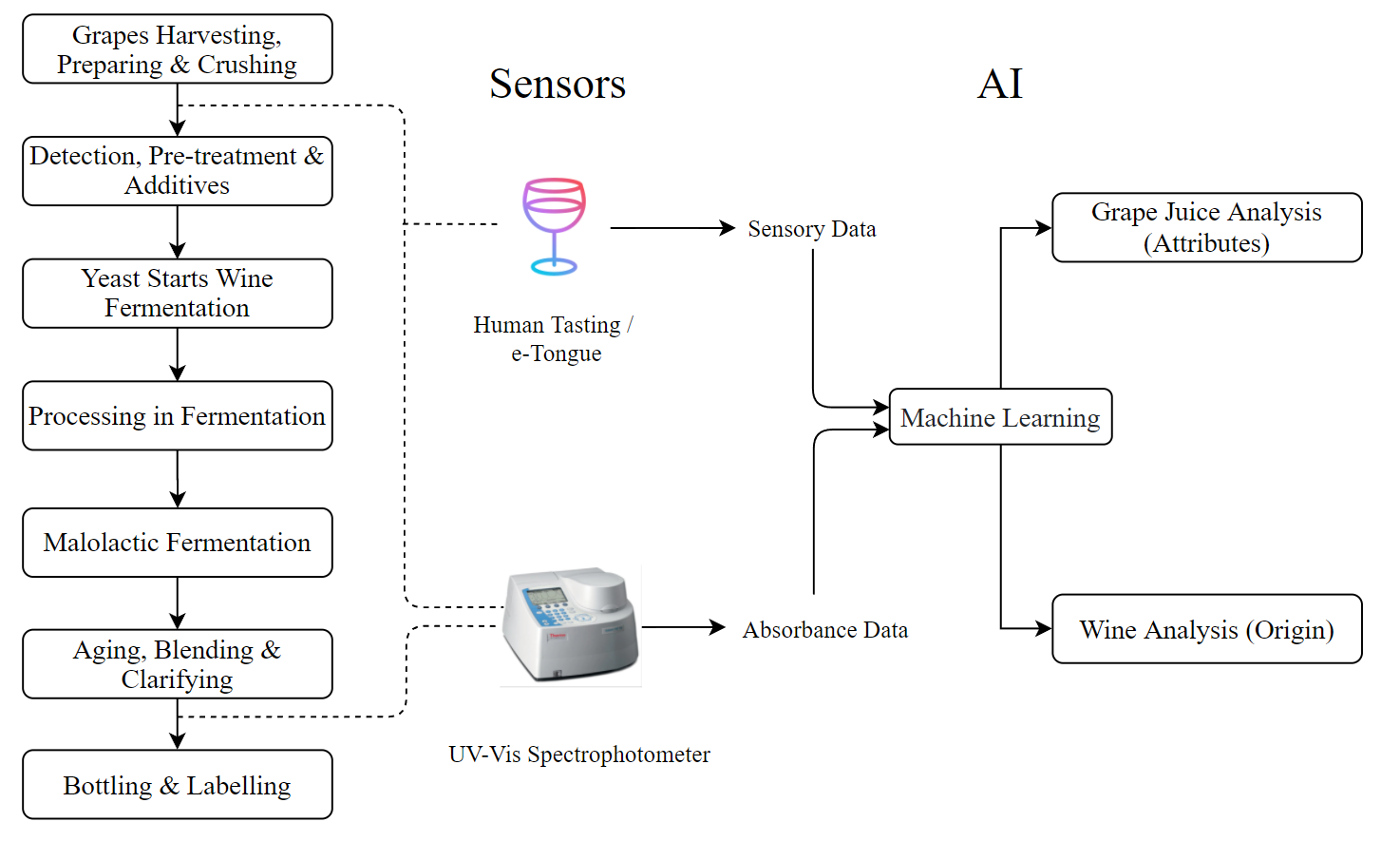}
\vskip -.5cm
\caption{Wine Process}
\label{wine_process}
\vskip -.75cm
\end{figure}

With the increasing demand for smart agriculture and production, wine analysis and new technologies for rapid analysis are developed. Due to wine containing multiple chemical components \cite{Valero2018}, manually executed analyses still take a large proportion of the wine production time. The concept of Smart Winery was introduced into the winemaking industry to reduce the costs associated with labour, uncertainty and errors, simplify and stabilize the winemaking process \cite{edselc.8505891676620190101,edssjs.DB1419FF20190101,edssjs.566BD07F20200101,edssjs.B0DB434020070101,edselc.2-52.0-8508699029920200701}. For example, Longo et al. explained the possibility and method of timely and objectively monitoring the grape juice to better control, monitor and optimize the winemaking process \cite{LONGO2021106810}. As illustrated by Figure \ref{wine_process}, The chemical composition during the process between grapes are harvested and crushed to make grape juice (GJ) will impact the final wine quality and flavor. For example, phenolics impact wine color, and palate attributes \cite{Blanco1998}. During this step, human tasting can be used to collect wine sensory data in different winemaking stages before bottling to label the samples' attributes. Sensory data is popular to be used as the attributes to analyze the wine's chemical composition as their representation. 

The traditional wine attributes analysis relies primarily on wine experts' tasting, which is very time-consuming and expensive. In addition, many factors, including physical conditions and residues in the mouth, affect human tastes, causing analysis results sometimes to be inaccurate. Therefore, the wine industry is always searching for a more automated, less costly, and accurate way to extract the sensory data of wine, which triggers the combination of IoT and AI in this field to offer a better solution.

Some sensors have been employed to automate the wine attributes collection process. For example, the electronic nose (E-nose) can monitor the evolution of desirable chemicals in place of human tasters. However, the data collected by these sensors is yet to be commercialized \cite{Valero2018} because complex sensors like E-nose require preprocessing the raw data to digitalize them for analysis. Therefore, the collected data can only be analyzed by professionals and objective mechanisms to effectively determine and identify the composition and quality of wine products. Compared with E-nose, ultraviolet-visible (UV-Vis) spectroscopy is a more standard method to assess parameters throughout the winemaking process with a UV-Vis scanning spectrophotometer. On the other hand, the development of artificial intelligence (AI), especially machine learning (ML), has been successfully applied in most aspects of wine research and winemaking \cite{12112785620160101}, \cite{Gonzalez_2019}, \cite{edseee.453716420081101}, \cite{edselc.8505891676620190101}, \cite{Thakkar2016AHPAM}. However, most of the research is not comprehensive enough to combine with sensors to identify the particular attributes of the wine. 

This paper comprehends two tasks, with both datasets provided by the Tasmanian Institute of Agriculture (TIA), Launceston, Tasmania (Australia). Task I investigates the relationship between UV-Vis and grape juice's key sensory features (astringency, bitterness, and herbaceous). In task I (i.e., regression task), we aimed to predict the data of the astringency, bitterness, and herbaceous characteristics of grape juice with different treatment through UV-Vis absorbance unit values, which were collected by UV-Vis scanning spectrophotometer. The astringency, bitterness, and herbaceous labels for each sample were collected by human tasting. Tasters are constituted by a number of panelists who received professional tasting training. The purpose of task II (i.e., classification task) is to classify the region and vineyard of the grape juice samples through basic juice parameters such as pH, total soluble solids (TSS), total acidity (TA), harvest type, and UV-Vis absorbance unit values. Both Tasks adopt the same set of ML algorithms - Random Forest (RF), Support Vector Machine (SVM), Deep Neural Networks (DNN, 1-3 layers), Convolutional Neural Network (CNN, 1 dimensional), Long Short-Term Memory (LSTM) and Bidirectional LSTM. However, the evaluation metrics for algorithms are different for two tasks - Task I as the regression task uses Mean Absolute Error (MAE), Root Mean Square Error (RMSE), and Explained Variance Score (EVS); Task II as the classification task uses Accuracy and F1 scores and tests them by the leave-one-out method. 

This paper finds an efficient method that combines with  ML approaches to identify grape juice's sensory characteristics by establishing the links between them and Spectroscopy. The sensory characteristics of grape juice under process are one of the keys that decide the wine's quality and category. The automation of sensory characteristics promotes the development of the smart winery industry by significantly saving time and cost. Spectral data combined with some basic chemical parameters are also adopted by the ML method to identify the wine's geographical origin, which is significant for its pricing and grading, to detect its authenticity. The wine industry welcomes new explorations and technologies that bring efficiency and reliability to production and quality.

\section{Related Work}

Sensory data is previously collected from human tasting and electronic sensors. In human tasting, a number of volunteers are recruited and offered a training. During this period, they acquire the ability to taste and score wine samples’ quality and various sensory parameters \cite{Maria_2019,edselc.2-52.0-8508699029920200701,edssjs.DB1419FF20190101}. Some researchers make efforts to replace human taste with machines\cite{edselc.8505891676620190101,edsdoj.26b1ac2e6ca6455aa232c1bef22a32c020161001}. For example, \cite{edselc.8505891676620190101} used a bionic electronic nose (E-Nose) to collect the sensory data. Their results gained support and confidence in the reliability of machine sensors relative to human tasting. \cite{edsdoj.26b1ac2e6ca6455aa232c1bef22a32c020161001} introduced a series of E-Noses and Tongues for the wine tasting to be developed as the trends. 
For origin prediction, \cite{edssjs.566BD07F20200101} used chemical elements to identify wines’ origin. They determined 13 elements (e.g., Al, Cd \& Co), which effectively distinguish origin and authenticity. The study of \cite{edssjs.B0DB434020070101} aimed to use spectroscopy in the visible and wavelength near-infrared (NIR) regions to do the non-destructive measurement of wines and predict the chemical compositions (e.g., ethanol, Free/Total SO\textsubscript{2} \& pH). Their research proved that spectroscopy is feasible and effective to predict the basic chemical composition of wine.

Thousands of companies and organisations take advantage of ML's ability to help with decision-making and prediction to save the labour, time, and cost of analysis \cite{12112785620160101,Gonzalez_2019}. 
In past years, SVM has proved its good robustness and performance for high-dimensional space cases \cite{Maria_2019}, and it is widely used to deal with the classification or regression problem of wine quality prediction \cite{edseee.453716420081101,edselc.8505891676620190101,Thakkar2016AHPAM}. \cite{James_2018} tried to adopt SVM to process quality and price prediction, but they failed to achieve an ideal result. The main goal of \cite{edseee.870201720190201}'s team was to compare two feature selection methods, Simulated Annealing (SA) and Genetic Algorithm (GA) . In \cite{edselc.8505891676620190101}, SVM was shown to achieve the best performance in vintages and fermentation processes prediction while Neural Networks was the best in production areas and varietals predictionBackpropagation.  RF has also been commonly used and achieved good results in various studies \cite{edselc.8505891676620190101,Thakkar2016AHPAM,Er_Y_Atasoy}. For example, \cite{Er_Y_Atasoy} stated that RF is better than K-nearest Neighbourhood and SVM in their wine quality prediction. \cite{edseee.900911120180701} relabelled the Red Wine Dataset (UCI) data to make it a binary classification problem (good/bad) and compared Logistic Regression, and RF's effect combined with 10-fold cross-validation . The RF improved the classification by $8\%$ to $85\%$ compared to Logistic Regression's $77\%$. In addition, \cite{edseee.910409520200101}'s team used the same dataset to compare the performance of SVM, RF and Naïve Bayes, and the experimental results obtained were $67.25\%$, $65.83\%$ and $55.91\%$ respectively. However, compared to SVM and RF,  \cite{10.1007/978-981-15-2780-7_120} stated Artificial Neural Networks (ANN), after appropriate tuning, can get a better prediction accuracy than SVM in small datasets. \cite{edselc.2-52.0-8508699029920200701} established three ANN regression models. Model 1 used 100 wavelength's raw absorbance values (1596-2396 nm) to predict the sensory data of Pinot Noir. Model 2 and 3 used weather and water balance data to predict sensory data and wine colour (CIELab, RGB with CMYK), respectively. 

Although machine learning has been applied to wine data in several applications, this paper is the first work to comprehensively study different ML models, including shallow learning and deep learning approaches,  for different tasks (regression and classification). The use of spectral information extracted from wine and grape juices for prediction is promising for real-life applications.

\section{The Color of GJ/WINE}
Colour that developed in the pressing process (pressing of the grapes to get juice) is one of the essential factors for grape juice and wine, not only because it influences the customer’s purchase decision, but also because it can be used to identify varieties \cite{3174421220191117}, estimate the phenolic content in wine \cite{3223549620200330}, judge wine quality \cite{UTAS.2765020180101}, and detect adulteration and frauds \cite{S030881462030691920200901}.
 
Monitoring of colour development is vital to winemakers to assist in controlling the consistency and quality of wine during production. Typically, winemakers may establish control or reference value by archiving their product data into a database that can be readily compared between different batches of products. Then in production, real-time comparison of the wine colour conducted by analysts may grasp the process of wine manufacturing and ensure the normal fermentation of wine products.

\begin{wrapfigure}{r}{0.45\textwidth}
	\vskip -1.cm
	\centering
	\includegraphics[width=0.45\textwidth]{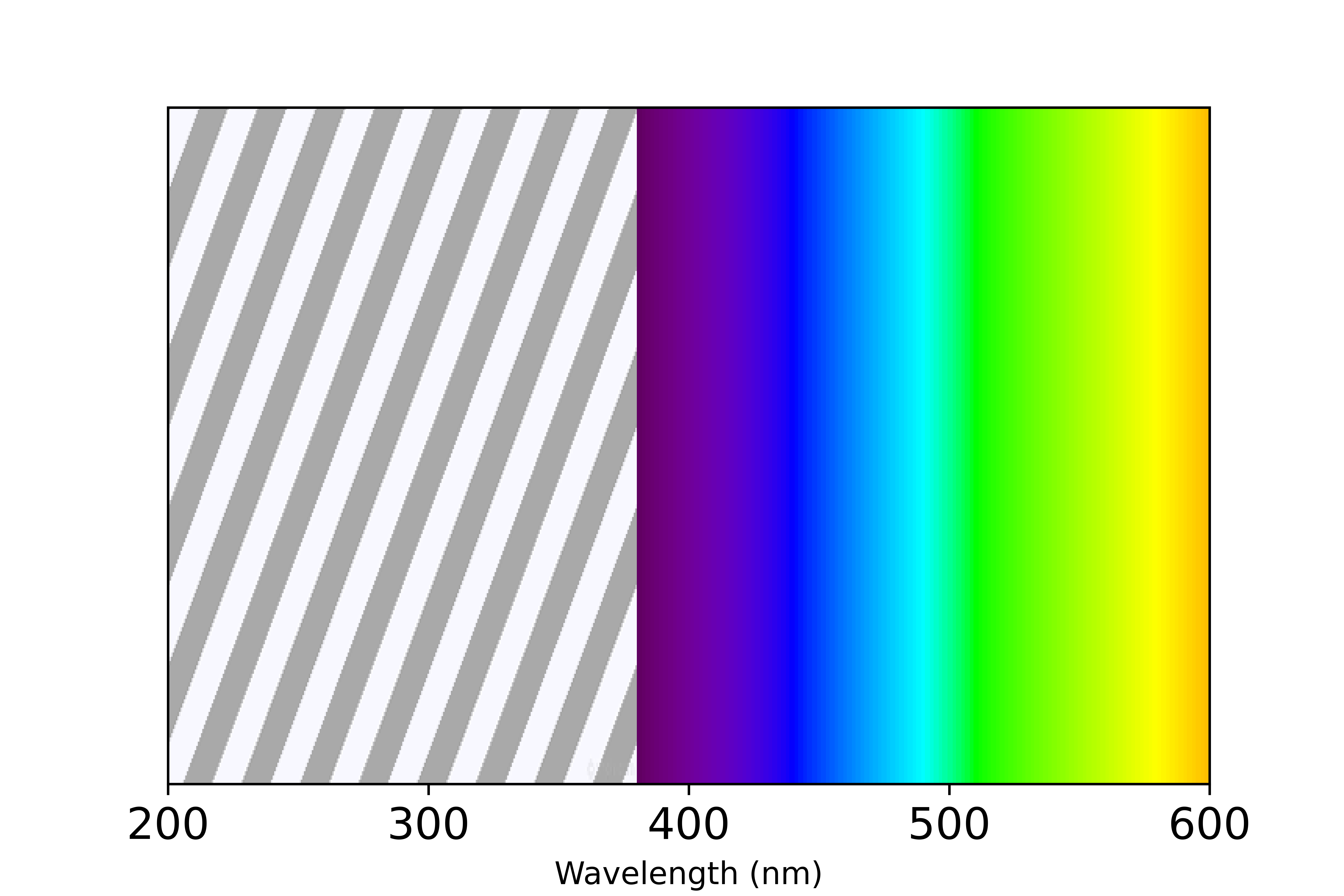}
	\caption{Absorbance Spectrum}
	\label{spectrum}
	\vskip -0.8cm
\end{wrapfigure}

Figure \ref{spectrum} shows the contrast spectrogram of the irradiation wavelength from 200 nm to 600 nm. Generally, the light below 380nm is ultraviolet light and cannot be seen by human eyes, so we used grey and white stripes instead. From 380 to 600 nm, we can see violet, blue, green, yellow, and orange from left to right.

Spectroscopy or spectral fingerprinting technology has been used widely in the wine industry because it is a mature, accessible, and reliable technology to analyse wine quality through its colour. UV-Vis spectroscopy (the ultraviolet region is 180–390 nm and visible region is 390–780 nm) is one of the most commonly used branches in this technology. Other types of spectral fingerprinting technology include near-infrared (NIR), infrared (IR), mid-infrared (MIR), nuclear magnetic resonance (NMR),  and mass (MS) \cite{UTAS.2765020180101}. UV-Vis spectroscopy has been used to quantify compounds in the food and beverage industries, such as polyphenolics in red wine products \cite{S030881461400265920140901}. The most significant advantage of this technology is that it is fast and straightforward to apply and requires less technical training \cite{S030881460500337720060101}. The spectral fingerprint relates to the light absorbance in the UV-Vis range of the electromagnetic spectrum according to Beer-Lambert law formula: 
\begin{equation}
A = \epsilon lc
\end{equation}

Because the molar attenuation coefficient ($\epsilon$) of the solution is the same, when the concentration of the solution (c) and the optical path length (l) is fixed, the absorbance (A) of the same solution’s samples should be close or the same. Therefore, if we measure the absorbance of wine in different process steps beforehand and record these data into a table, then we can determine the quality of GJ / wine through the absorbance. Considering the reliability and effectiveness of spectroscopy, some wine analysis professionals believed that more spectral databases of a variety of wines should be developed \cite{3174421220191117}. We choose to train the ML models on absorbance data of GJ for the same reason.

\section{Grape Juice Tasting}
\subsection{Data Collection}
The absorbance data of grape Juice is used to train the regression model to predict the values of three sensory attributes (Astringency, Bitterness, and Herbaceous), and its collecting process will be explained in the next section - Juice (or Grape) Origin Prediction. This section mainly introduces the process we collected the labeling data (the three sensory attributes) of each sample. A panel of 11 panelists (six males, five females; 23-59 years), the post-graduate students or staff of TIA, was convened. Each panelist had completed a minimum of five training sessions over the two weeks prior to the formal sensory analysis. The protocols and training for the descriptive analysis (DA) that we adopt are based on a research paper published in the journal "Food Research International" \cite{LONGO2018561}. Briefly, a generic DA protocol was applied \cite{lawless2010sensory}, with one-hour training sessions for all descriptors: 'sweetness', 'astringency', 'bitterness', 'acidity', and 'herbaceous'. A sample of each grape juice treatment (n = 7) was provided during the training sessions. Panelists practiced the evaluation of intensity rating for all attributes. 'High' and 'low' intensity standards were obtained using different concentrations of reference standards in water or juice (i.e., 15 and 150 g/L fructose for 'sweetness,' 0.1 and 1 g/L quinine sulfate for 'bitterness,' 1.5 and 15 g/L tartaric acids for 'acidity,' 0.1 and 1 g/L tannin for 'astringency,' one-quarter or one cup of fresh grass for 'herbaceous'). While we were mainly interested in the 'astringency' and 'bitterness' descriptors for the Chardonnay grape juice, as associated with the presence of phenolics, other taste and flavor attributes were included to avoid a 'dumping' effect (i.e., dumping occurs when panelists are limited to responding to only one attribute at the time).
As per standard formal DA, panelists were seated and served seven pre-poured covered glasses containing juice samples in randomized order (Williams block design), with each glass was assigned with a 3-digit code. Individual spittoons, citric acid-pectin solution, and unsalted crackers were also served for a mouth rinse and to reduce palate fatigue. The panelists used ballot charts and pencils to rate the intensity of each descriptor from 0 to 9 (0 = absent; 9 = very intense). Three formal DA sessions were conducted to assess each treatment three times formally. Social science ethics approval for collecting tasting data was obtained from the University of Tasmania's Research Integrity and Ethics Unit (Ref No: H0018377). The 11 judges scored each GJ replicate. Total phenolic content and the UV-Vis absorbance were measured (from 200 to 600 nm) for each replicate.

For evaluation, we use Mean Absolute Error (MSE) and Root Mean Square Error (RMS). In addition, Explained Variance Score (EVS) should also be applied to measure the model’s total variance. Its formula is as follow:

\begin{equation}
\textit{explained\_variance(y,ŷ)}= 1 -\frac{\textit{Var\{y-ŷ\}}}{\textit{Var\{y\}}}
\end{equation}

\textup{y} is the correct (true) label value, \textup{ŷ} is the predicted value of the model, and Var is the Variance. The best EVS can be 1.0, while the low value represents poor performance.

\subsection{Prediction Results}

\begin{table}
\vskip -0.5cm
\centering
\caption{Astringency, Bitterness, and Herbaceous estimation using Machine Learning models.}\label{lab:taste}
\resizebox{0.8\textwidth}{!}{%
\begin{tabular}{l|ccc||ccc||ccc}
\hline
Model& \multicolumn{3}{|c|}{Astringency Est.}& \multicolumn{3}{|c|}{Bitterness Est.}& \multicolumn{3}{|c}{Herbaceous Est.}\\
 & MAE$\downarrow$ & RMSE$\downarrow$ & EVS$\uparrow$ &MAE$\downarrow$ & RMSE$\downarrow$ & EVS$\uparrow$ & MAE$\downarrow$ & RMSE$\downarrow$ & EVS$\uparrow$\\
\hline
\hline
 SVR  & 1.287 & 1.759& 0.336 & 0.792 & 1.250 & 0.526 & 1.402 & 1.769 & 0.356\\
 \hline
 RF &1.671 &2.062 &0.086 & 1.180 &1.533 &0.268 & 1.658&2.019 & 0.149\\
 \hline
 DNN.1 &1.518 & 1.900 & 0.226 & 0.992 &1.400 &0.390 & 1.506 & 1.836&0.296\\
 \hline
 DNN.2 & 1.660 & 2.037 & 0.109 & 1.206 & 1.564 & 0.239 & 1.621 & 1.998 & 0.166\\
 \hline
 DNN.3 & 1.687 & 2.052 & 0.095 & 1.272 & 1.646 & 0.157 & 1.650 & 2.048 & 0.124\\
 \hline
 1D-CNN &1.787 &2.167 &-0.009 & 1.433 &1.800 &-0.009 & 1.767& 2.198& -0.009\\
 \hline
 LSTM &1.565 &1.951 &0.182 & 1.104 &1.480 &0.318 & 1.546&1.893 &0.251\\
 \hline
 bi-LSTM &1.607 &1.977 &0.160 & 1.036 &1.437 &0.357 & 1.543 &1.932 &0.220 \\
 \hline
 \hline
\end{tabular}
}
\vskip -.6cm

\end{table}

In Task 1, eight ML models have been tested on the attributes of the grape juice samples, which are Astringency (Table \ref{lab:taste}: Astringency Est.), Bitterness (Table \ref{lab:taste}: Bitterness Est.) and Herbaceous (Table \ref{lab:taste}: Herbaceous Est.). For astringency estimation, SVR had the highest performance in the prediction of astringency - the scores of MAE and RMSE were the smallest, and the EVS score was the largest. DNN (1 layer) and LSTM were the second and third best models. (SVR) also had the best performance in Bitterness prediction - achieved the best scores in all three evaluation metrics. DNN (1 layer) and bi-LSTM followed. In Herbaceous prediction (Table \ref{lab:taste}: Herbaceous Est.), the best model rankings were SVR, DNN (1 layer), and LSTM. The gap between 
LSTM and bi-LSTM was not obvious because the latter's MAE score was slightly better than the former. Overall, SVR (linear kernel) had the best performance in attributes prediction tasks. Compared with other tested models, it shows certain robustness. The second-best algorithm identified is DNN (1 layer). However, it should be noted that the size of the dataset limited the prediction performance of each model, making prediction tasks difficult. The more detailed and more extensive datasets will be more helpful for robust prediction.

\section{Juice (or Grape) Origin Prediction}
\subsection{Data Collection}
\begin{figure*}[ht]
	\vskip -.8cm
	\centering
	\begin{subfigure}{0.4\textwidth}
		\centering
		\includegraphics[width=0.9\textwidth]{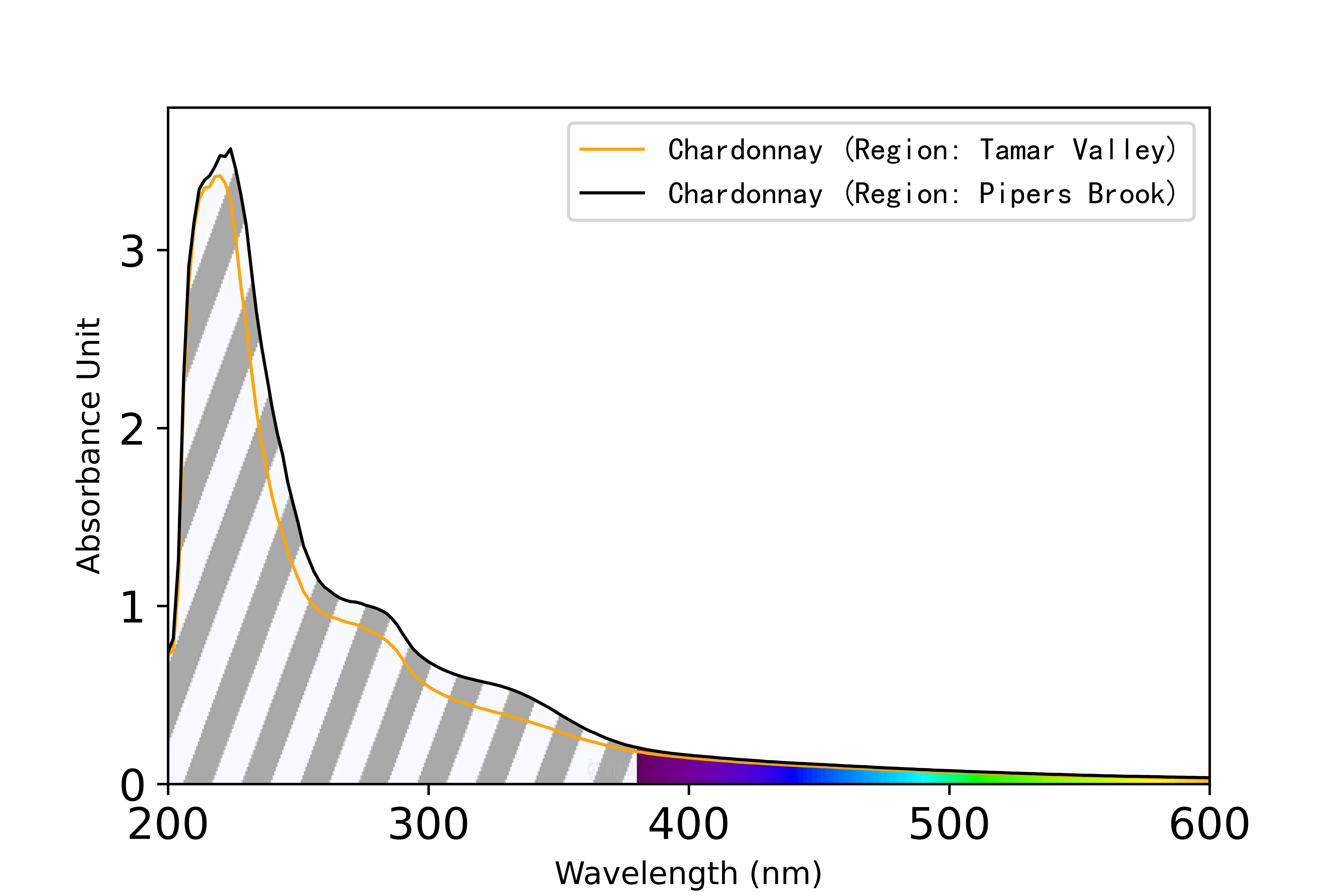}
		\caption{Different Region Samples}
		\label{fig:region_color}
	\end{subfigure}
	\begin{subfigure}{0.4\textwidth}
		\centering
		\includegraphics[width=0.9\textwidth]{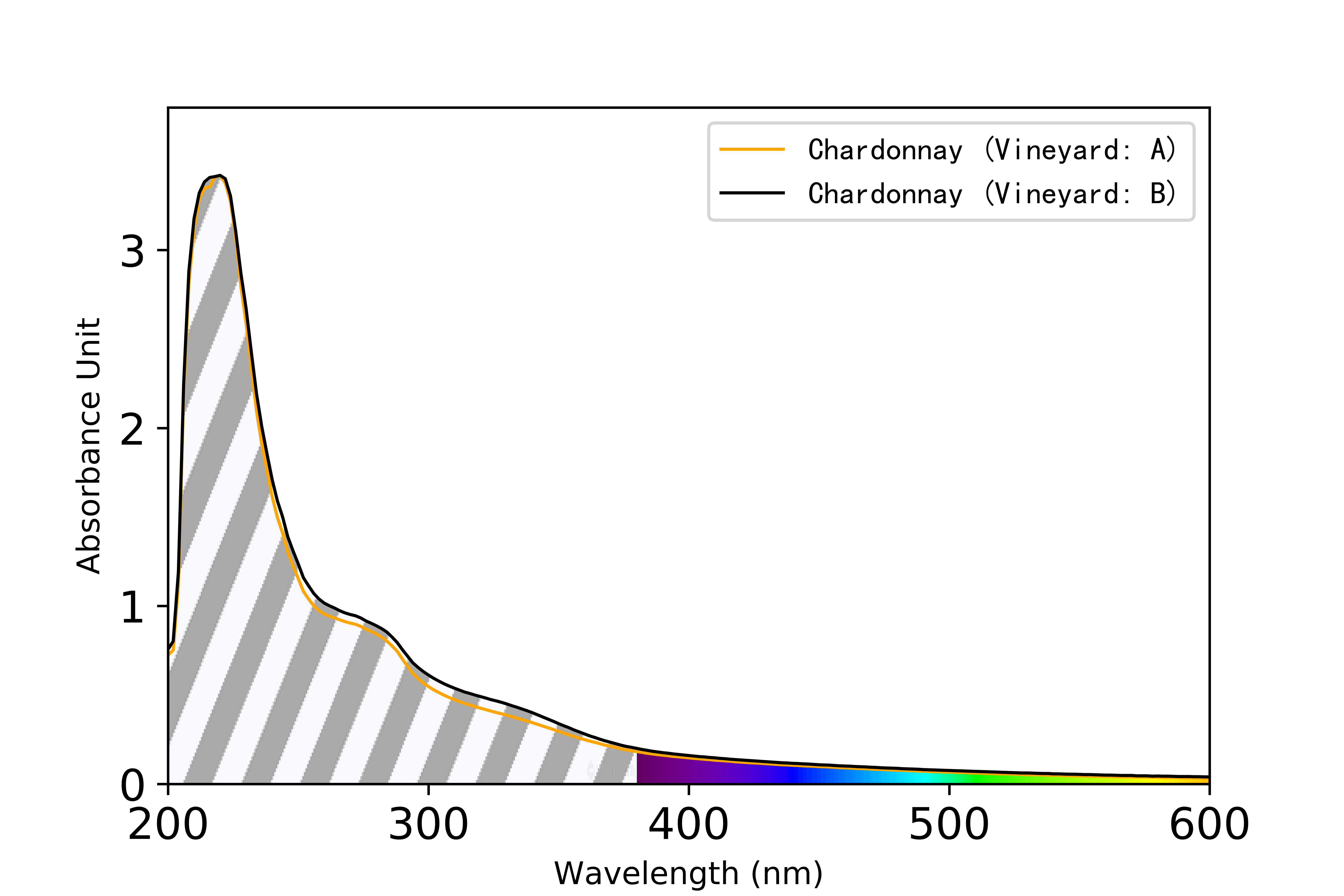}
		\centering
		\caption{Different Vineyards }
		\label{fig:vinyard_color}
	\end{subfigure}
	\begin{subfigure}{0.4\textwidth}
		\centering
		\includegraphics[width=0.9\textwidth]{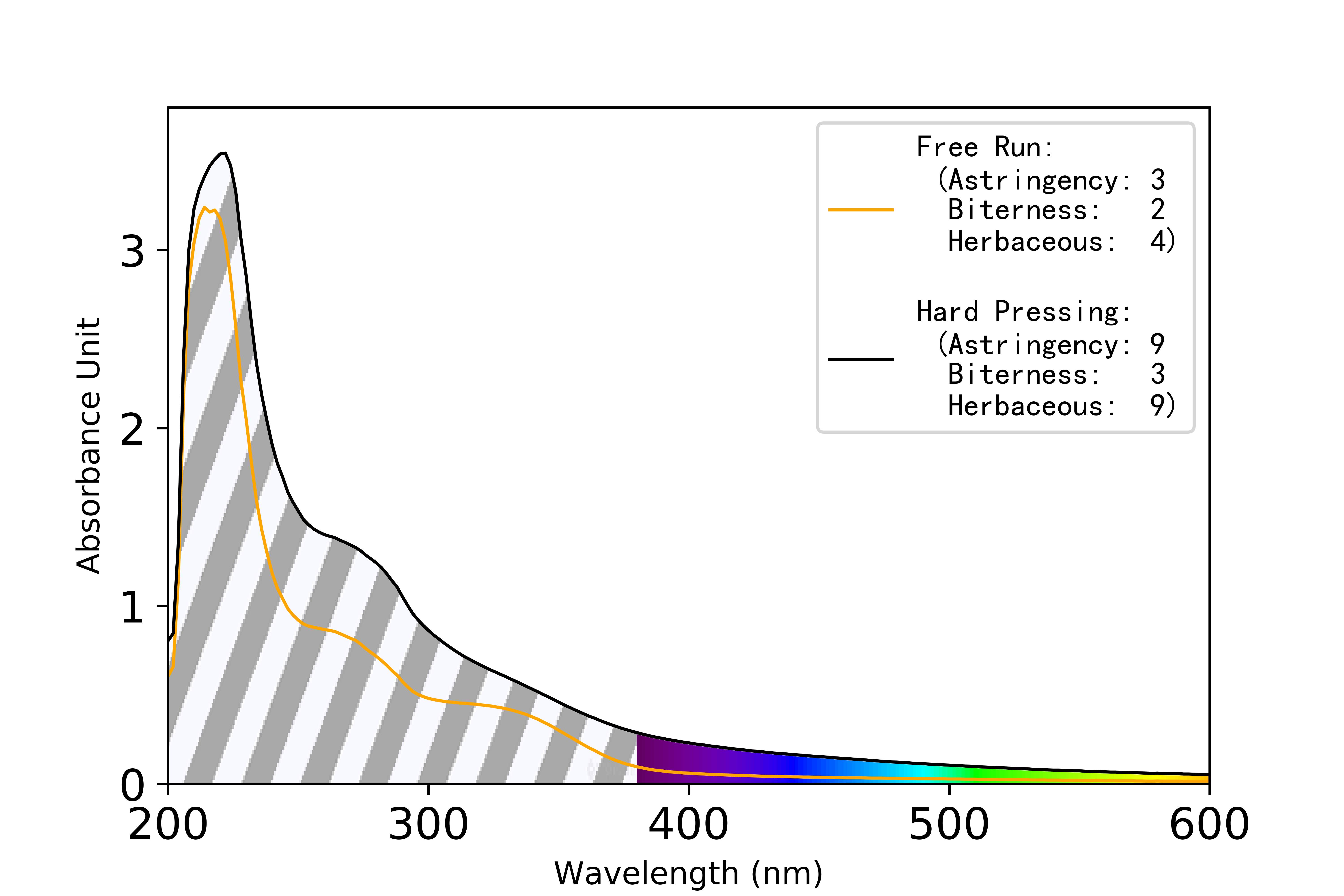}
		\caption{Free Run and Hard Pressing}
		\label{fig:FR_HP}
	\end{subfigure}
	\begin{subfigure}{0.4\textwidth}
		\centering
		\includegraphics[width=0.9\textwidth]{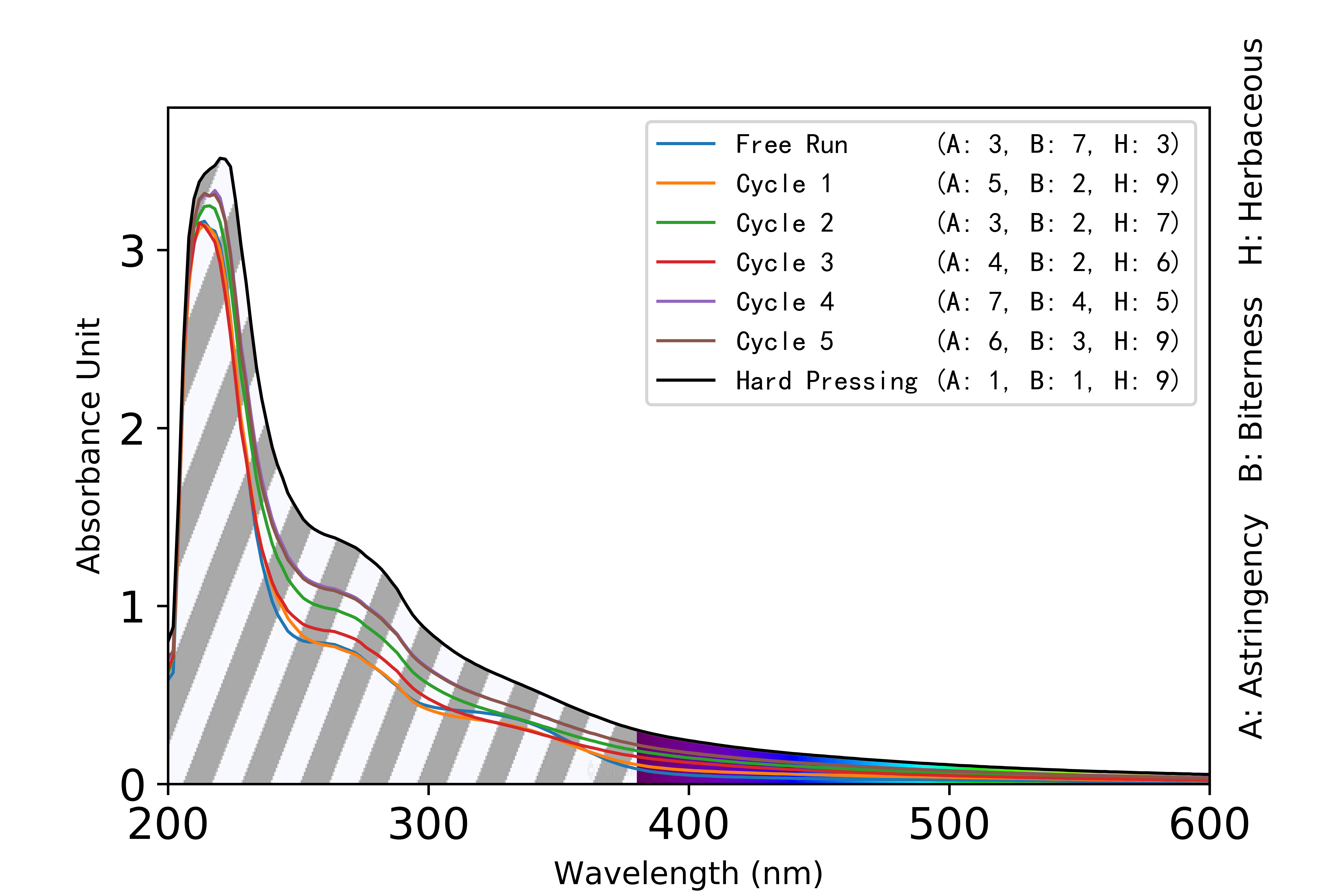}
		\caption{All Treatments}
		\label{fig:AllTreatments}
	\end{subfigure}
        \vskip -.2cm
	\caption{the Absorbance Data of Samples}
	        \vskip -.8cm
\end{figure*}

In this task, juice samples from Chardonnay and Pinot noir grapes were kept at 4$^{\circ}$C overnight. Then, they were centrifuged using a 5804 Eppendorf (Hamburg, Germany) for 15 min (at 3350 radial centrifugal force), diluted at a ratio of 1 to 5 with 1 M HCl (Merck, Darmstadt, Germany), and incubated in darkness at an ambient temperature (22$^{\circ}$C) for 1 hour. Next, we scan the samples with a Genesys 10S UV-Vis spectrophotometer (Thermo Scientific, Waltham, MA, USA) and record the absorbance unit values (contained the spectral phenolic fingerprint) every 2 nm from 200 to 600 nm. Because the samples were placed in disposable 10 mm quartz cuvettes (Brand-GMHB, Wetheim, Germany), specific wavelengths (below 250 nm) were discounted. In the dataset, each sample has the following categories: variety, vineyard, region, the block of its region, harvest type (e.g., by hand or by machine), the replicate number, TSS, pH value, and TA and absorbance unit values (from 200 to 600 nm, in 2nm increments). We visualize the data to show the details of the absorbance data. We can see that, in Figure \ref{fig:region_color}, the absorbance units status of two Chardonnay grape juice samples are from different regions, which are Tamar Valley and Pipers Brook (Tasmania, Australia), with peaks at around 220 nm. Figure \ref{fig:vinyard_color} shows two Chardonnay samples from different vineyards (we called them A and B) in the same region (Tamar Valley). We can see from the absorbance unit values of the Chardonnay sample from A (black line) that almost all its values are higher than the sample from B. Figure \ref{fig:FR_HP} shows two Grape juice samples from different points during the grape pressing process, Free Run and Hard Pressing. We can see they are very different in UV-Vis absorbance unit values. The black one (Free Run)'s values are much higher than the Hard Pressing sample. In terms of attributes, their bitterness scores are similar (Free Run is 2 and Hard Pressing is 3); their Astringency and Herbaceous values are very different (Free Run has 3, 4 respectively, and both values of Hard Pressing are 9). Figure \ref{fig:AllTreatments} shows the details about all treatments of the grape juice, the black line (Hard Pressing) on the top; Free run (blue), and Cycle 1 (Orange) at the bottom of all lines. The details of their related attributes are shown in the upper right corner, among which 'A' represents Astringency, 'B' represents Bitterness, and 'H' represents Herbaceous.

\subsection{Prediction Results}

\begin{table*}[h]
\begin{center}
\caption{Region and Vineyard Prediction}\label{Region_and_Vineyard}
\resizebox{0.8\textwidth}{!}{%
\begin{tabular}{c|cc||cc|||cc||cc}
\hline
& \multicolumn{4}{c|||}{Leave-one-juice-out} & \multicolumn{4}{|c}{Leave-one-sample-out} 
\\
& \multicolumn{2}{|c|}{Region} & \multicolumn{2}{|c|||}{Vineyard}& \multicolumn{2}{|c|}{Region} & \multicolumn{2}{|c}{Vineyard}
\\
Model   & Accuracy & F1 & Accuracy & F1 & Accuracy & F1 & Accuracy & F1 \\
\hline
\hline
 SVM  &0.559 &0.556 &0.172 &0.144 & 0.914 & 0.913 &0.925 &0.926 \\
 \hline
 RF &0.624 &0.618 &0.161 &0.139  &0.796 &0.785 &0.710 &0.704 \\
 \hline
 DNN.1 &0.677 & 0.673 &0.108 &0.100 &0.742 &0.733 &0.613 &0.594\\
 \hline
 DNN.2 &0.602 &0.591 &0.129 &0.118 &0.796 &0.791 &0.699 &0.695\\
 \hline
 DNN.3 &0.591 &0.591 &0.118 &0.118 &0.785 &0.769 &0.667 &0.655\\
 \hline
 1D-CNN &0.430 &0.431 &0.065 &0.058 &0.828 & 0.826 &0.796 &0.778 \\
 \hline
 LSTM &0.581 &0.586 & 0.172 & 0.166 & 0.806 &0.800 &0.860  &0.861\\
 \hline
 bi-LSTM &0.505 & 0.504 &0.108 &0.096 &0.839 &0.837 &0.849 &0.848 \\
 \hline
 \hline
\end{tabular}%
}
\end{center}

\vskip -.5cm
\end{table*}

Table \ref{Region_and_Vineyard} is the results of task II. In leave-one-juice-out, we can see that all models have similar performance. DNN (1-layer) is relatively better than other models in terms of Region prediction. LSTM has relatively better performance in terms of Vineyard prediction, but their advantage is not significant because based on the leave-one-juice-out method (It classified the group by the type of samples, and the dataset is divided into 31 parts in total), the types of samples are relatively rare. Especially in vineyard prediction, because the number of the vineyard class is more than others, The problem that there is no group class in the training set corresponding to the test set also appears. Therefore, the result of vineyard prediction through leave-one-juice-out is not good and effective. Through leave-one-juice-out, the model is difficult to be trained enough. In contrast, leave-one-sample-out method is more effective and accurate. In this method, the dataset was divided into 93 parts, using one part as the test set each time. Most algorithms have ideal results around 80$\%$, among which SVM (linear kernel) has the best performance (more than 90$\%$). It has the highest accuracy and F1 score in both Region and Vineyard predictions. Compared with other algorithms, all the DNN methods do not provide satisfactory results, which may be due to the small size of the dataset impacting on performance. As in Table \ref{Region_and_Vineyard}, SVM shows the strongest performance. Especially through the leave-one-sample-out method, which can be used to predict both Region and Vineyard.
\begin{table*}[h]
\vskip -0.5cm

\begin{center}
\caption{Wavelength Importance Ranking}\label{WavelengthImportance}
\begin{tabular}{ccccccccccc}
& \multicolumn{2}{c}{Astringency}& \multicolumn{2}{c}{Bitterness}& \multicolumn{2}{c}{Herbaceous}& \multicolumn{2}{c}{Region}& \multicolumn{2}{c}{Vineyard} \\

 No. & RF & SVR & RF & SVR & RF & SVR & RF & SVC & RF & SVC \\

1 & 210 & 204 & 204 & 204 & 216 & 204 & 402 & 232 & 206 & 206 \\
2 & 348 & 220 & 206 & 206 & 222 & 206 & 206 & 230 & 204 & 260 \\
3 & 350 & 200 & 244 & 220 & 204 & 212 & 420 & 210 & 218 & 262 \\
4 & 204 & 208 & 330 & 222 & 206 & 222 & 414 & 228 & 212 & 264 \\
5 & 208 & 216 & 238 & 218 & 322 & 220 & 202 & 246 & 208 & 266 \\

\end{tabular}%
\end{center}

\vskip -0.5cm
\end{table*}
\section{Wavelength Importance}
\begin{figure*}[h]
\vskip -.5cm
	\centering
	\begin{subfigure}{\textwidth}
		\centering
		\includegraphics[width=\textwidth]{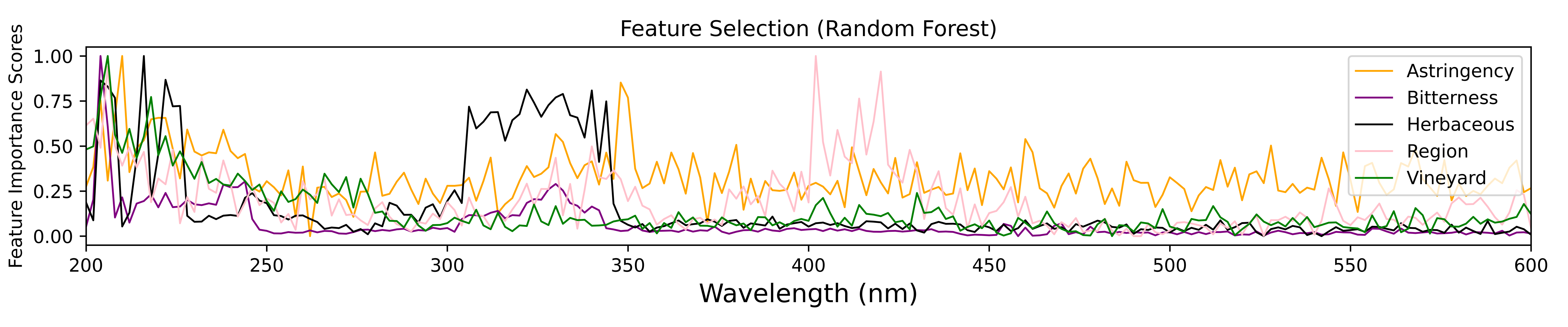}
  \vskip -.3cm
		\caption{Wavelength Importance Scores (Random Forest)}
		\label{fig:Feature_Importance_Random_Forest}
	\end{subfigure}
	\begin{subfigure}{\textwidth}
		\centering
		\includegraphics[width=\textwidth]{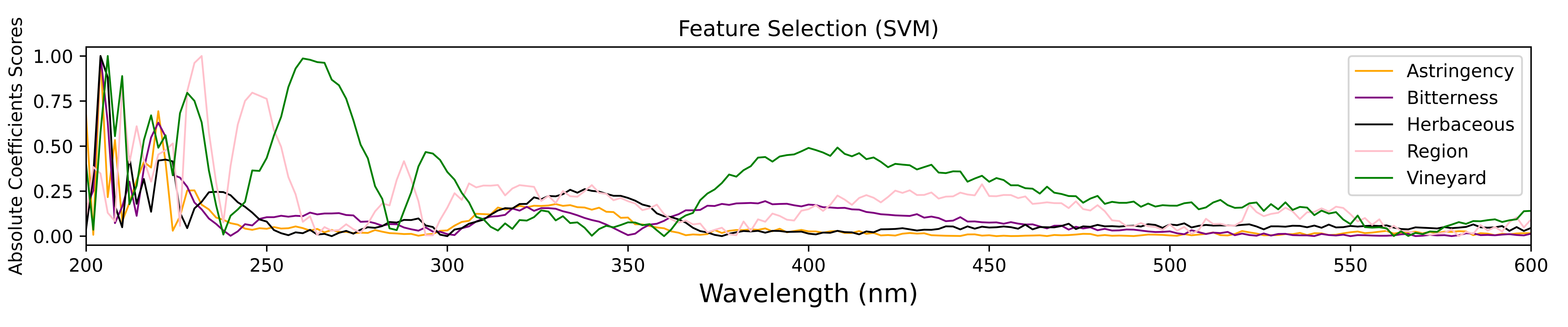}
		\centering
        \vskip -.3cm
		\caption{Absolute Coefficients Scores (SVM) }
		\label{fig:Feature_Importance_SVM}
	\end{subfigure}
 \vskip -.1cm
	\caption{the Visualization of Wavelength Importance}
	 \vskip -1cm
\end{figure*}

Table \ref{WavelengthImportance} is the wavelength importance ranking of GJ attributes and origin prediction through RF and SVM (linear kernel). We selected the top 5 most important wavelengths of each object. To better show the importance of each wavelength, we visualized their scores in RF (Figure \ref{fig:Feature_Importance_Random_Forest}) and SVM (Figure \ref{fig:Feature_Importance_SVM}). The peak of each line is the most important wavelength of the object. In RF, the value of the Y-axis is the feature importance score, which is similar to the absolute coefficients scores that also represent the importance as the Y-axis value in SVM. The two groups of importance values were normalized (from 0 to 1). For example, the wavelength of attribute bitterness in both RF and SVM reaches the peak (the most important) at 204nm. According to the results (Table \ref{WavelengthImportance}), in RF, the relatively more important wavelengths for attribute prediction are 204 and 222nm. In SVM, 216, 220, 204, and 222nm are more important. Thus, 204 and 222nm are the top two wavelengths to predict the sensory data. RF selected 200, 202, 206, and 208nm for GJ origin prediction, while SVM selected 210nm as the most important in both region and vineyard prediction. From Table III, we can see that the highest-scoring wavelengths are from 200 to 400nm, with a small part being outside 400nm. Thus the most important wavelengths are from 200 to 420 nm will be helpful for the GJ/wine analysis (attributes and origin prediction). In Figure \ref{fig:Feature_Importance_Random_Forest} and Figure \ref{fig:Feature_Importance_SVM}, from 200 to 250nm, at the front of the wavelength range, the peaks and troughs show that prediction of this range will be unstable so that we would choose the more stable and relatively high range (250-420nm).

\section{Conclusions}
This paper effectively combined ML with IoT technique using UV-Vis spectroscopy to predict GJ's sensory attributes and the origin of grapes (region and vineyard). The experimental results can provide guidance and reference for industrial production regarding wine analysis. Based on the results, SVM has the best efficiency among serval algorithms in these tasks, whether attributes prediction or origin prediction. Especially in origin prediction, it achieved more than $91\%$ in accuracy and F1 score, proving its' practical value in production or quality inspection. By analyzing the important wavelengths, SVM has also provided a particular wavelength range that could be used for validation or select sensors. Therefore, this paper brings new ideas to wine producers and other beverage or food production areas by providing solutions to automatically obtain the wine's sensory and origins data for wine quality, categories, or authenticity analysis. In the future, we will look for further integration of ML technology and wine analysis to assist with winemaking, such as vineyard weather prediction or wine variety prediction. We believe that smart wineries that combine machine learning, IoT, and big data will be the future trend in the wine production industry because they will further improve the quality of wine and reduce costs and risks for wineries.

\end{document}